# Exploring the Use of ChatGPT for a Systematic Literature Review: a Design-Based Research

Qian Huang, Lee Kuan Yew Centre for Innovative Cities, Singapore University of Technology and Design, Singapore
Email: qian_huang@sutd.edu.sg

Qiyun Wang, National Institute of Education, Nanyang Technological University, Singapore
Email: qiyun.wang@nie.edu.sg

**Abstract**

ChatGPT has been used in several educational contexts, including learning, teaching and research. It also has potential to conduct the systematic literature review (SLR). However, there are limited empirical studies on how to use ChatGPT in conducting a SLR. Based on a SLR published, this study used ChatGPT to conduct a SLR of the same 33 papers in a design-based approach, to see what the differences are by comparing the reviews' results, and to answer: To what extent can ChatGPT conduct SLR? What strategies can human researchers utilize to structure prompts for ChatGPT that enhance the reliability and validity of a SLR? This study found that ChatGPT could conduct a SLR. It needs detailed and accurate prompts to analyze the literature. It also has limitations. Guiding principles are summarised from this study for researchers to follow when they need to conduct SLRs using ChatGPT.

## Introduction

A systematic literature review (SLR) is "a research methodology to collect, identify, and critically analyze the available research studies (e.g., papers, conference proceedings, books, dissertations) through a systematic procedure" (Carrera-Rivera, Ochoa, Larrinaga, & Lasa, 2022, p2). It advances evidence-based knowledge by synthesizing common themes and trends in existing studies over a period of time (Kitchenham et al., 2009; Moher et al., 2009) and identify research gaps for future studies (Gough, Oliver, & Thomas, 2017).

However, doing an SLR is a highly labour intensive and time-consuming task in framing clear research questions and developing effective strategies for literature search and analysis (Cohen, Manion, & Morrison, 2018; Sliverman, 2021). The recent generative AI (GAI) tools like ChatGPT have promising affordances in reading papers and summarizing common themes. It seems that they have great potential in helping researchers do the SLR.

To the best of our knowledge, no empirical study has been conducted to explore how to use GAI like ChatGPT to do an SLR. Some existing conceptual papers have presented the idea of using ChatGPT to do literature reviews but no one has explored and published papers on how to do it in practice (Haman & Školník, 2023; Waseen, Ahmad, Liang, & Fahmideh, 2023). Therefore, this study was initiated to explore how to effectively use ChatGPT to do an SLR. More specifically, this study aims to answer the following research question:

- Can ChatGPT carry out a systematic literature review? If so, how?

## ChatGPT

ChatGPT, developed by OpenAI, represents a significant advancement in natural language processing and artificial intelligence. Its origins trace back to the introduction of the GPT (Generative Pre-trained Transformer) architecture, with GPT-3, released in June 2020, being a notable predecessor that set the groundwork for large-scale language models. ChatGPT specifically was designed to improve interaction with AI, offering users the ability to engage in conversational dialogue, ask questions, receive answers, and even generate content across various formats and styles. It utilizes deep learning techniques to understand and generate human-like text based on the input it receives.

The transition from ChatGPT 3.5 to GPT-4.0 marked a pivotal evolution, introducing significant enhancements in understanding, context awareness, and generative abilities. ChatGPT 3.5 improved conversational interactions across various applications, while GPT-4.0, unveiled in early 2023, advanced these capabilities further, offering nuanced, contextually accurate responses and a closer approximation to human-like reasoning. This progression underscores OpenAI's dedication to refining AI technologies, aiming to make them more effective, creative, and user-friendly, thus expanding their utility in daily tasks and professional fields alike.

ChatGPT is a language generating tool. It is useful for generating a text description or improving writing (Wu et al., 2023). In addition to language learning, ChatGPT also has great potential in other aspects of education. In the existing literature, researchers have identified some potential benefits of using ChatGPT in education. For instance, it can design quizzes (Eysenbach, 2023, Jahic, Ebner & Schön, 2023); promote active learning because learners need to refine the answers ChatGPT generated based on their understanding (Shoufan, 2023); strengthen collaborative learning (Cotton, Cotton, & Shipway., 2023); promote personalized learning, especially for learners with disabilities (Cotton et al., 2023); provide immediate feedback and personlized learning (Kuhail et al., 2022); serve as writing assistant (Jahic et al., 2023); help with data analysis in research (Hwang & Chen, 2023, Kooli, 2023).

However, it also has some challenges. For instance, learners may be trained as being lack of the ability to generate original ideas and be critical (Arif, Munaf, & Ul-Haque, 2023; Choi, Lee, Ho., Kwok, & Lok, 2023; Kooli, 2023); learning process changed (Qi, Zhu, & Wu, 2023); it may cause misinterpretation in research (Kooli, 2023); Ethical issues, for instance, the authorship in academic writing (da Silva, 2023, Zhang & Zhen, 2023), plagiarism (Jahic et al., 2023). The existing literature has also made some suggestions for the use of ChatGPT in education, for instance, need the correct prompts to generate the answers (Thakur,

Devanshi & Thakur, 2023); develop learners' new digital skills and competencies (Kasneci et al., 2023).

## Methodology

As an exploratory study, this study aimed to investigate how to use ChatGPT to conduct an SLR for the purpose of generating equivalent results to the outcome of an original literature review (OLR) accomplished by the Wang and Huang (2023). In the OLR, 33 papers were systematically selected by following the PRISMA flowchart and deliberately analysed and synthesized. The findings were presented based on two research questions: 1) What challenges are involved in BSL that cause the low engagement of online learners? 2) What strategies are used to address the challenges and increase the engagement of online learners?

In the present literature review (PLR), the same papers were uploaded and analysed by ChatGPT-4.0 in the same way as in the OLR process. The two research questions were initially used as prompts to instruct GPT to generate review results. However, after testing with a few papers, we found that the generated result by GPT was unacceptable. For instance, GPT did not focus on the findings reported in the results section of each paper. But rather, it also reported points mentioned in the literature review section of a paper. To make the review findings more accurate, prompts were adjusted. After three rounds of major adjustment, ChatGPT could finally generate similar outcomes to those presented in the OLR.

As shown in Figure 1, in the OLR, two researchers read the 1st paper and discussed it. After reading 5 papers, they discussed the common challenges and strategies. Later, they discussed and summarized after the 12th and 33rd papers. Similarly, in this PLR, the researchers instructed ChatGPT to generate review results at the same points.

Table 1 shows an overview of the changes in each round. The following sections describe how each round was carried out, what prompts were used, how ChatGPT reacted to the prompts, and how the results were gradually improved.

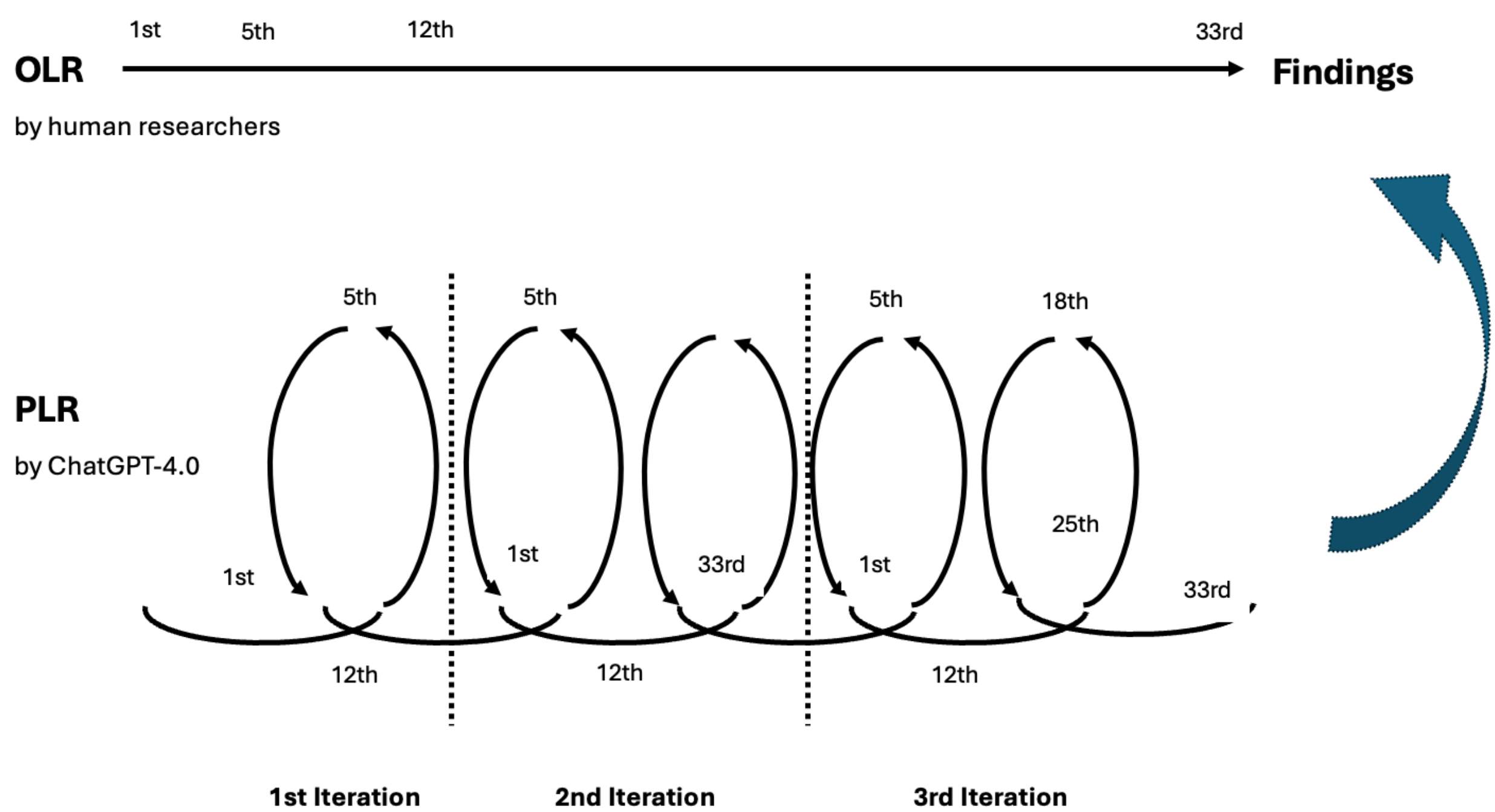

Figure 1. Iterative process of the study

**Table 1.** Overview of the DBR process

| | 1st Round | 2nd Round | 3rd Round |
|---|---|---|---|
| **Step 1: Read each paper and list the challenges and strategies** | Full paper | Finding/Results, Conclusion, Discussion by limiting page number | Finding/Results, Conclusion, Discussion by limiting page number |
| **(Number of paper read)** | 12 for trial | 33 | 33 |
| **Step 2: Summarize the common challenges and strategies and codify using given analytical framework** | None | Four times: 5th, 12th, 33rd | Six times: 5th, 12th, 18th, 25th, 33rd |
| **Step 3: Conduct an Overview** | None | Done | Done |

## The Iterative Processes

In this section, the implementation process, results, and improvement in each round are described to illustrate how ChatGPT gradually improved review results.

### Round 1

<u>Process</u>

The literature analysis process with ChatGPT was initially identical to that of OLR. A researcher uploaded the 1st paper onto ChatGPT 4.0 via its Plugin named "Ai PDF". The initial prompt used were the two research questions as follows:

> *(Prompt) Please read the paper and answer the following two questions: What challenges are involved in Blended Synchronous Learning that cause the low engagement of online learners? What strategies are used to address the challenges and increase the engagement of online learners?*

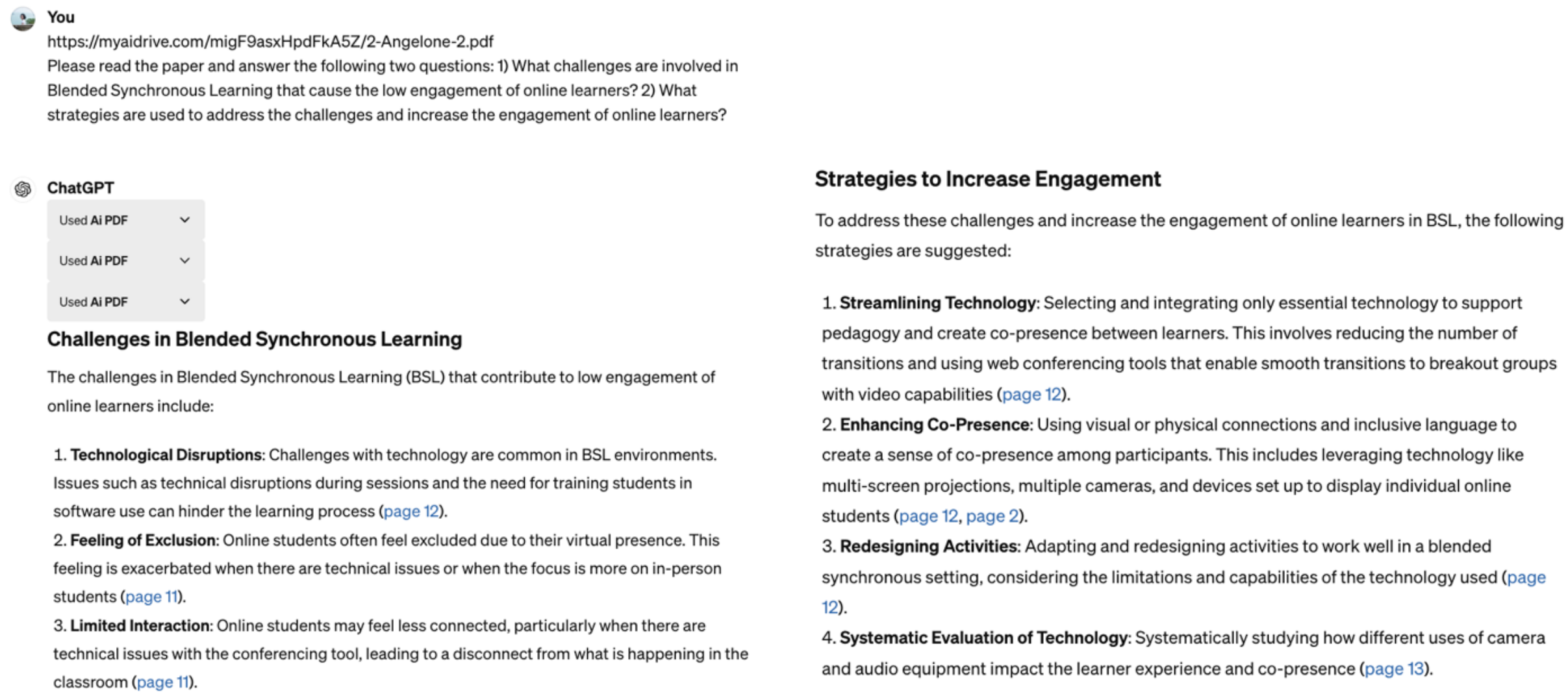

Figure 3. The result of reading the 1st reviewed paper

As shown in Figure 3, ChatGPT could summarise strategies, indicating it had high potential to perform an SLR. Subsequently, ChatGPT was further tasked to continue reading four additional papers. The researcher instructed GPT to summarise the common challenges and strategies of the first 5 papers by using the following prompt:

> *(Prompt) For challenges, any common themes from these 5 papers? and for strategies, any common themes?*

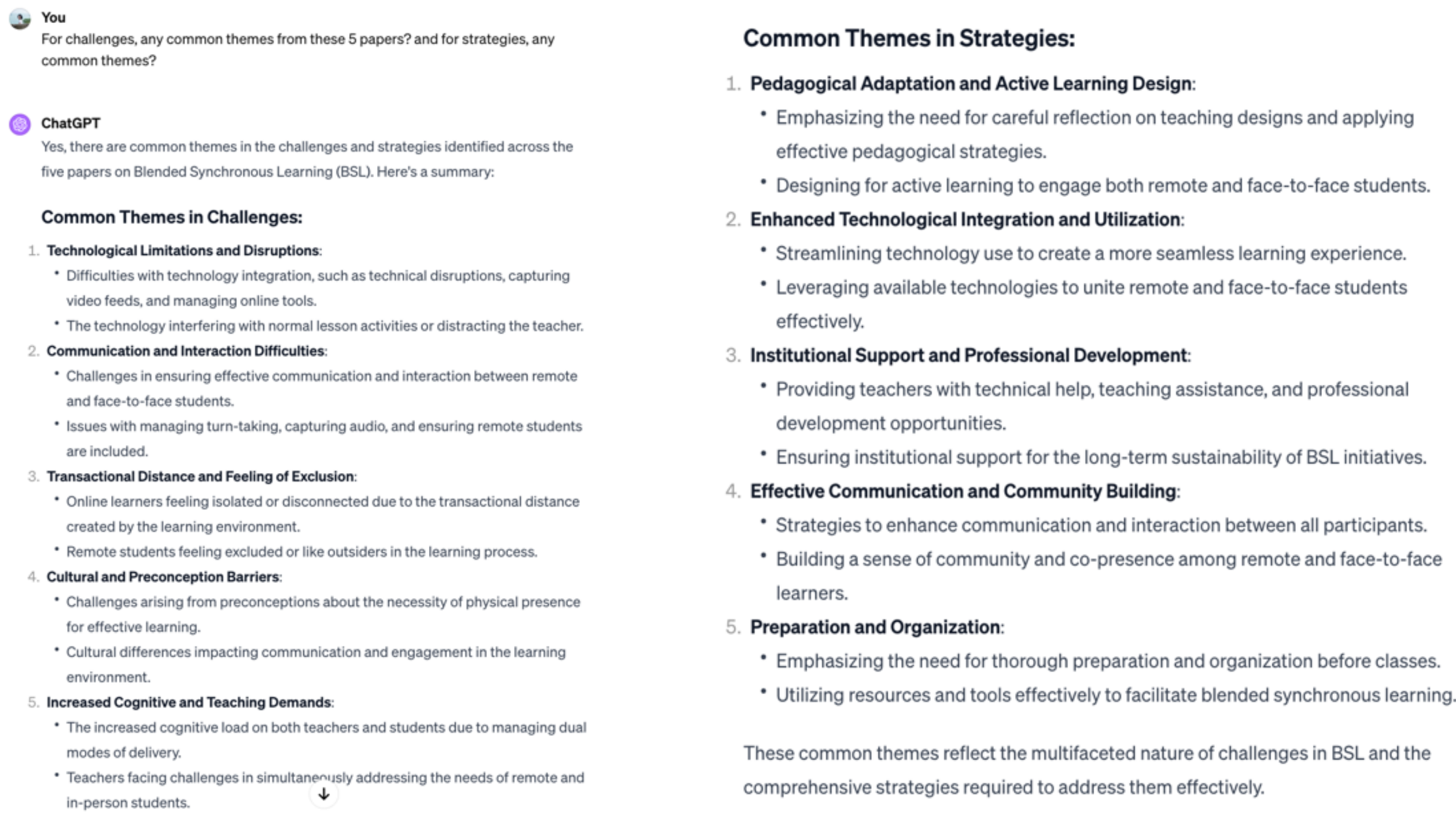
You
For challenges, any common themes from these 5 papers? and for strategies, any common themes?

ChatGPT
Yes, there are common themes in the challenges and strategies identified across the five papers on Blended Synchronous Learning (BSL). Here's a summary:

### Common Themes in Challenges:

1. **Technological Limitations and Disruptions**:
   - Difficulties with technology integration, such as technical disruptions, capturing video feeds, and managing online tools.
   - The technology interfering with normal lesson activities or distracting the teacher.
2. **Communication and Interaction Difficulties**:
   - Challenges in ensuring effective communication and interaction between remote and face-to-face students.
   - Issues with managing turn-taking, capturing audio, and ensuring remote students are included.
3. **Transactional Distance and Feeling of Exclusion**:
   - Online learners feeling isolated or disconnected due to the transactional distance created by the learning environment.
   - Remote students feeling excluded or like outsiders in the learning process.
4. **Cultural and Preconception Barriers**:
   - Challenges arising from preconceptions about the necessity of physical presence for effective learning.
   - Cultural differences impacting communication and engagement in the learning environment.
5. **Increased Cognitive and Teaching Demands**:
   - The increased cognitive load on both teachers and students due to managing dual modes of delivery.
   - Teachers facing challenges in simultaneously addressing the needs of remote and in-person students.

### Common Themes in Strategies:

1. **Pedagogical Adaptation and Active Learning Design**:
   - Emphasizing the need for careful reflection on teaching designs and applying effective pedagogical strategies.
   - Designing for active learning to engage both remote and face-to-face students.
2. **Enhanced Technological Integration and Utilization**:
   - Streamlining technology use to create a more seamless learning experience.
   - Leveraging available technologies to unite remote and face-to-face students effectively.
3. **Institutional Support and Professional Development**:
   - Providing teachers with technical help, teaching assistance, and professional development opportunities.
   - Ensuring institutional support for the long-term sustainability of BSL initiatives.
4. **Effective Communication and Community Building**:
   - Strategies to enhance communication and interaction between all participants.
   - Building a sense of community and co-presence among remote and face-to-face learners.
5. **Preparation and Organization**:
   - Emphasizing the need for thorough preparation and organization before classes.
   - Utilizing resources and tools effectively to facilitate blended synchronous learning.

These common themes reflect the multifaceted nature of challenges in BSL and the comprehensive strategies required to address them effectively.

Figure 4. The common challenges and strategies after reading the first 5 papers

As shown in Figure 4, the generated results confirmed that ChatGPT could do an SLR. However, it seemed that the generated results from the 3rd paper (as illustrated in Figure 5) were not accurate. For instance, one challenge summarized by GPT was: *"Unmet Expectations: Early studies evaluating the effectiveness of educational use of VC indicated that it had not yet met participants' expectations."* Clearly, this challenge was obtained from the "literature" section but not from the results of the paper.

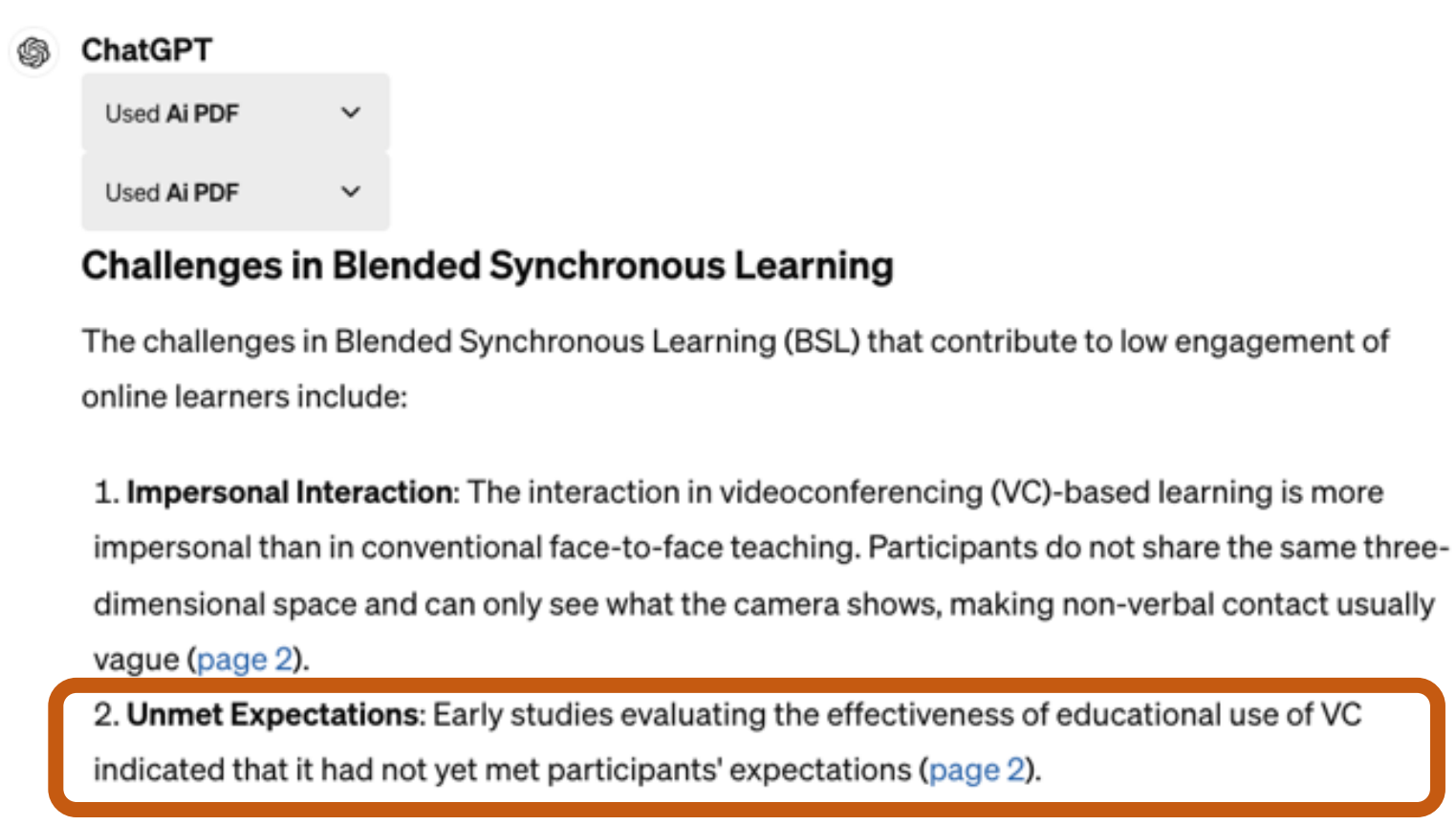
ChatGPT
Used **Ai PDF**
Used **Ai PDF**

### Challenges in Blended Synchronous Learning

The challenges in Blended Synchronous Learning (BSL) that contribute to low engagement of online learners include:

1. **Impersonal Interaction**: The interaction in videoconferencing (VC)-based learning is more impersonal than in conventional face-to-face teaching. Participants do not share the same three-dimensional space and can only see what the camera shows, making non-verbal contact usually vague (page 2).
2. **Unmet Expectations**: Early studies evaluating the effectiveness of educational use of VC indicated that it had not yet met participants' expectations (page 2).

Figure 5. The challenges of the 3rd paper reported by ChatGPT

To further confirm the review accuracy, we further instructed ChatGPT to generate common challenges and strategies after analysing 12 papers by creating a table. The next section presents the results.

## Results and Analysis

The generated results of the 12 papers are shown in Figure 6. It displays the challenges and strategies. It seemed that the generated table was well structured. However, the results had some issues.

| Paper | Challenges in BSL | Strategies to Increase Engagement |
|---|---|---|
| **Paper 1** | - Impersonal Interaction<br>- Unmet Expectations | - Combining VC with Web Environments<br>- Reinforcement of Social Interaction<br>- Face-to-Face Collaboration<br>- Improvement of Evaluation Methodology |
| **Paper 2** | - Technological Disruptions<br>- Feeling of Exclusion<br>- Limited Interaction | - Streamlining Technology<br>- Enhancing Co-Presence<br>- Redesigning Activities<br>- Systematic Evaluation of Technology |
| **Paper 3** | - Technological Limitations<br>- Communication Challenges<br>- Increased Teaching Demands | - Preparation and Organization<br>- Leveraging Available Technologies<br>- Task Design and Pedagogy |
| **Paper 4** | - Preconceptions and Cultural Limitations<br>- Transactional Distance<br>- Cognitive Load | - Careful Reflection and Pedagogical Strategies<br>- Design for Active Learning<br>- Institutional Support |
| **Paper 5** | - Technological and Cultural Limitations<br>- Transactional Distance<br>- Cultural Communication Patterns | - Pedagogical Strategies and Workarounds<br>- Design for Active Learning<br>- Institutional Support |
| **Paper 6** | - Lower Levels of Relatedness<br>- Limited Interactions<br>- Technological Disconnect | - Enhancing Peer Relatedness<br>- Addressing Technological Challenges<br>- Improving Instructor Impact and Program Structure |
| **Paper 7** | - Emotional Experiences<br>- Differences in Emotional Activation<br>- Complexity Added by Technology | - Understanding Emotional Experiences<br>- Tailoring Approaches<br>- Managing the Complexity of Technology |
| **Paper 8** | - Technical Issues<br>- Internet Connectivity<br>- User Friendliness<br>- Social Isolation | - Optimizing Technical Design<br>- Improving Internet Connectivity<br>- Training and Familiarization<br>- Addressing Social Isolation<br>- Incorporating Online Lectures into Assessment |
| **Paper 9** | - Balancing Contributions<br>- Technology and Professional Development<br>- Audio Delays and Interruptions | - Creating Meaningful Interaction<br>- Employing Instructional Strategies<br>- All Students Participating in Virtual Environment |
| **Paper 10** | - Unequal Distribution of Attention<br>- Technology Barriers<br>- Cognitive Overload for Instructors | - Creating Active Learning Activities<br>- Appointing a Student to Monitor Text Chat<br>- Professional Development for Instructors |
| **Paper 11** | - Student Isolation and Engagement<br>- Reluctance to Participate in Forums<br>- Technology Barriers | - Synchronous Online Interaction<br>- Facilitating Interaction Between Students<br>- Managing Speaking Rights |
| **Paper 12** | - Videoconferencing Challenges<br>- Negative Impact on Learning<br>- Perceived Inequities in Learning Experiences | - Improving Videoconferencing Experience<br>- Enhancing Communication and Engagement<br>- Addressing Perceived Inequities |

Figure 6. The common challenges and strategies for 12 papers

They were not accurate. The results confirmed that ChatGPT captured related information from the entire document when it was reading the paper in the PDF document. They included the challenges and strategies mentioned in the “Literature” section. It was supposed to be from the “Results” section only.

They were presented broadly. In analysing common challenges and strategies, a specific analytical framework (learner-learner, learner-content, learner-instructor interaction) was initially used in the OLR process. In this round, however, GPT did not report findings based on a framework, and the results were loosely reported.

The results given were general without detailed information like page numbers. It was hard to identify where the common findings were derived. To address these issues, another round of analysis was conducted as presented in the following section.

**Round 2**

Revised Process

To address issue 1, prompts were adjusted by narrowing down to specific sections and page range. For instance, we used prompts like “Please read pages 5-14” to instruct GPT to focus on the “Finding/Results” section of the 1st paper.

> *(Revised Prompt) Please read pages 5-14, and answer two questions: What challenges are involved in Blended Synchronous Learning that cause the low engagement of online learners? What strategies are used to address the challenges and increase the engagement of online learners?*

To address issue 2, ChatGPT was prompted to analyse common challenges and strategies based on the interaction framework used in the OLR (Learner-learner, Learner-content, Learner-instructor).

> *(New Prompt) Can you please categorize challenges into 1) learner-instructor interactions; 2) learner-learner interactions; 3) learner-content interactions?*

To address issue 3, ChatGPT was prompted to report common challenges and strategies with page numbers.

Moreover, ChatGPT was tasked to generate an overview on educational contexts, research methods, and design frameworks, as what we did in the OLR. The analysis scope was limited to the "Introduction" and "Research Methodology" sections of each paper. The next section shows the results of the 2$^{nd}$ round.

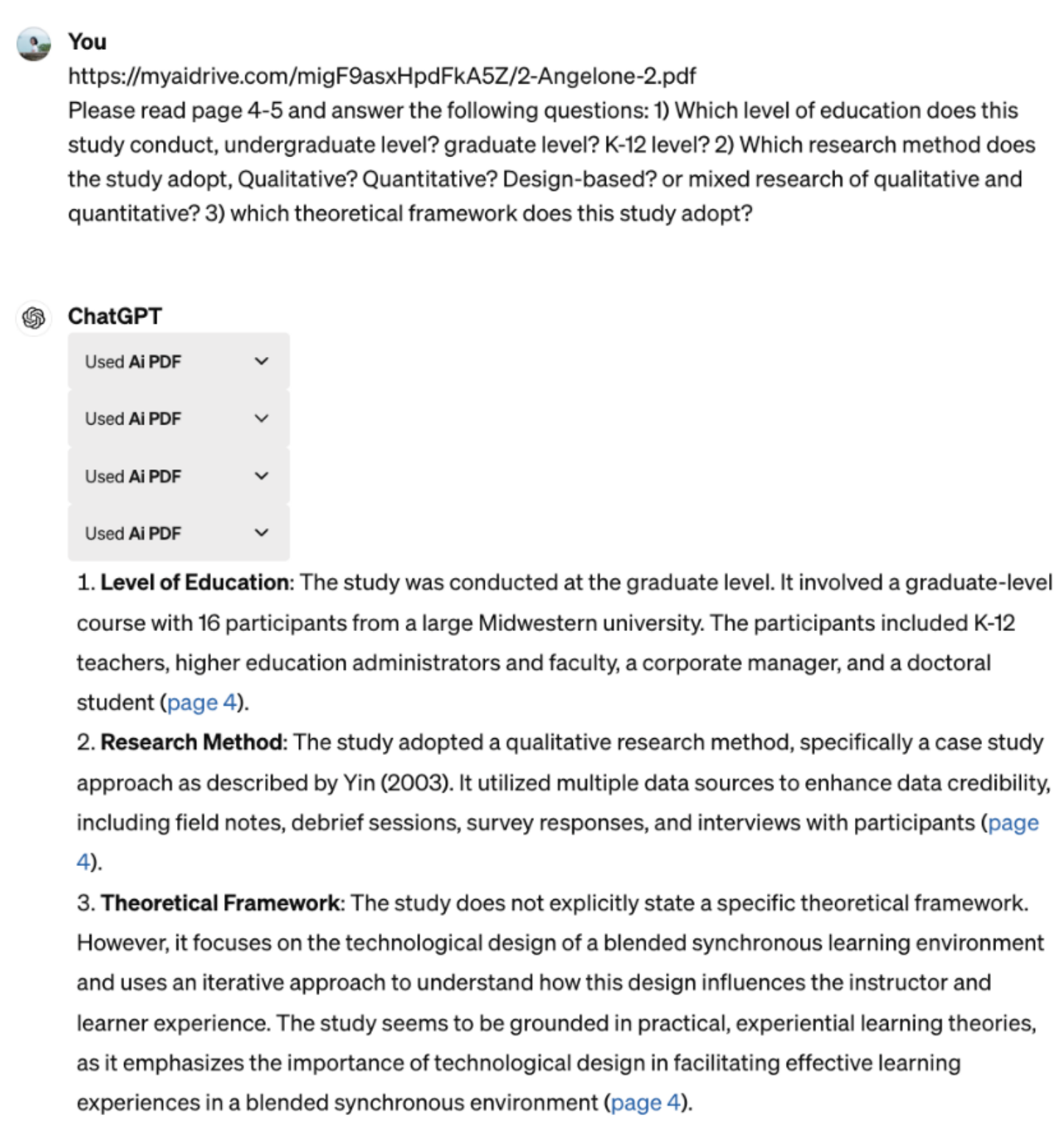

Figure 9. Prompts were given to do an Overview as OLR

Results and Analysis:

Figure 7 shows the results after narrowing down to specific sections and page range. Satisfactorily, GPT did not include any information from the “literature” section.

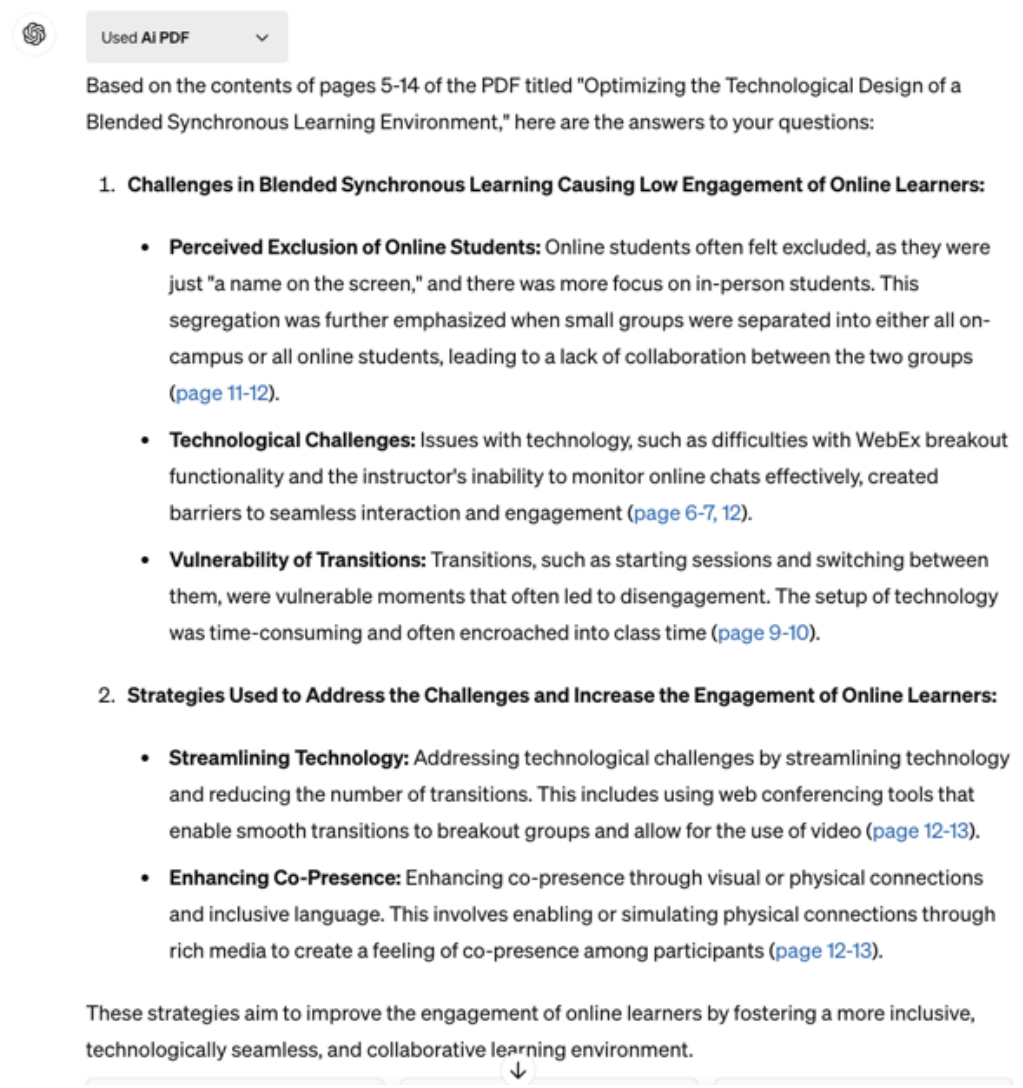

Used **Ai PDF**

Based on the contents of pages 5-14 of the PDF titled "Optimizing the Technological Design of a Blended Synchronous Learning Environment," here are the answers to your questions:

1. **Challenges in Blended Synchronous Learning Causing Low Engagement of Online Learners:**
   - **Perceived Exclusion of Online Students:** Online students often felt excluded, as they were just "a name on the screen," and there was more focus on in-person students. This segregation was further emphasized when small groups were separated into either all on-campus or all online students, leading to a lack of collaboration between the two groups (page 11-12).
   - **Technological Challenges:** Issues with technology, such as difficulties with WebEx breakout functionality and the instructor's inability to monitor online chats effectively, created barriers to seamless interaction and engagement (page 6-7, 12).
   - **Vulnerability of Transitions:** Transitions, such as starting sessions and switching between them, were vulnerable moments that often led to disengagement. The setup of technology was time-consuming and often encroached into class time (page 9-10).
2. **Strategies Used to Address the Challenges and Increase the Engagement of Online Learners:**
   - **Streamlining Technology:** Addressing technological challenges by streamlining technology and reducing the number of transitions. This includes using web conferencing tools that enable smooth transitions to breakout groups and allow for the use of video (page 12-13).
   - **Enhancing Co-Presence:** Enhancing co-presence through visual or physical connections and inclusive language. This involves enabling or simulating physical connections through rich media to create a feeling of co-presence among participants (page 12-13).

These strategies aim to improve the engagement of online learners by fostering a more inclusive, technologically seamless, and collaborative learning environment.

Figure 7. Prompts were adjusted by narrowing down to specific sections and page range

For the 2nd issue, GPT was prompts to analyze the common challenges and strategies based on the interaction framework. It seemed ChatGPT could analyse and categorize challenges and strategies based on the given framework, as illustrated in Figure 8.

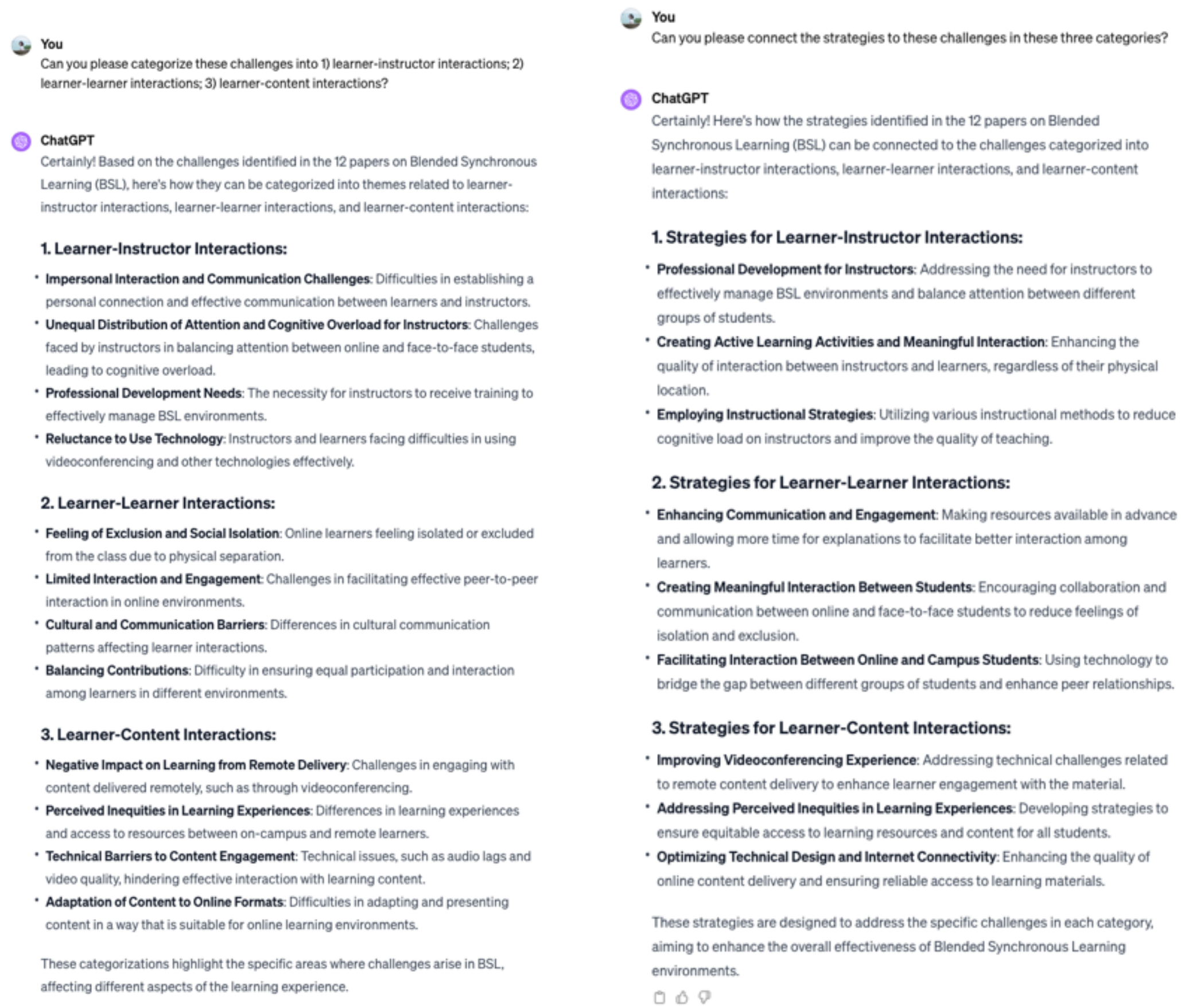

Figure 8. Analyzing the common challenges and strategies based on the iteration framework

For the 3rd issue, we found that ChatGPT could not identify page numbers for all papers by stating "Not explicitly discussed". In addition, the page numbers given were the ones of each PDF file starting from page 1, not the actual ones labelled in the journals.

In addition, we found that after reading many papers in one time, ChatGPT generated very limited common challenges and strategies as shown in Figure 9. For example, ChatGPT generated the challenges in learner-instructor interactions: increased cognitive load for instructors, balancing attentions, technological proficiency. In each theme, however, it only listed 1-2 papers.  We decided to add additional checkpoints (e.g., between the 12th and 33rd papers) to see if ChatGPT could generate richer results.

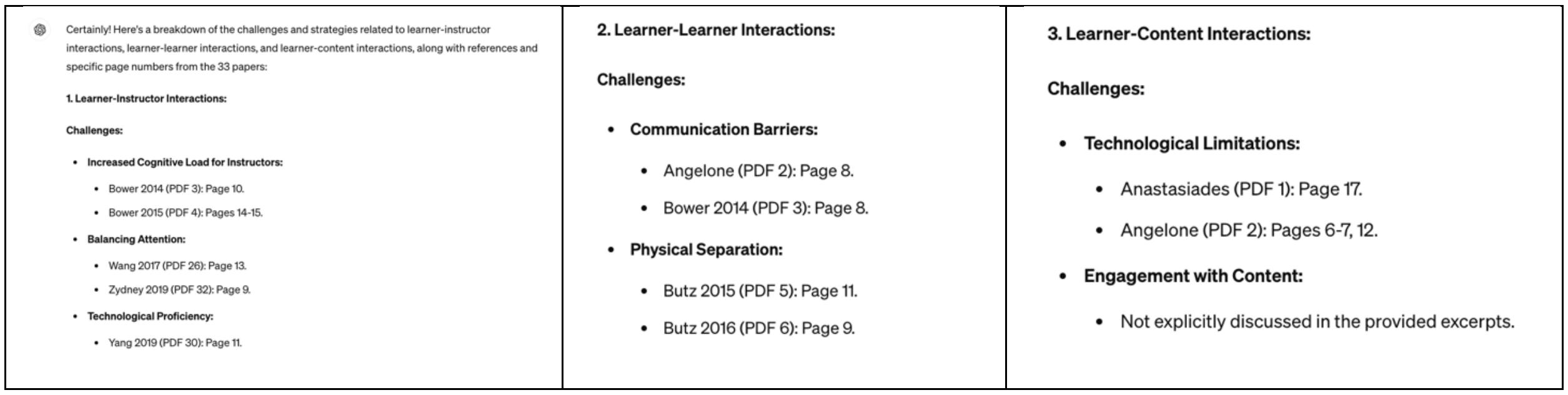

Figure 9. ChatGPT's results across 33 papers in 2nd round of iteration

In terms of the generated overview, some items (e.g., graduate level; a mix of graduate and undergraduate levels) were highly consistent with that of the OLR. However, some items (e.g., K-12, DBR, Framework) had big differences as shown in Table 2. In particular, ChatGPT could not identify which paper adopted the design-based research method. In addition, it could not identify accurately which theoretical framework each study adopted.

**Table 2.** Comparison of the Overview

| Category | Description (by Human Researchers) | ChatGPT |
|---|---|---|
| **Educational context** | - Higher education: 29<br>○ Graduate: 15<br>○ Undergraduate: 8<br>○ A mix of undergraduate and graduate: 1<br>○ Not identified: 4<br>- K-12: 4 | - Higher education: 23<br>○ Graduate: 15<br>○ Undergraduate: 7<br>○ A mix of undergraduate and graduate: 1<br>○ Not identified: 0<br>- K-12: 10<br>- |
| **Research method** | - Qualitative: 16<br>- Mixed methods: 11<br>- Design-based research: 8<br>- Quantitative: 6 | - Qualitative: 15<br>- Mixed methods: 13<br>- Design-based research: 1<br>- Quantitative: 4 |
| **Design framework** | - CoI: 8<br>- Transactional distance: 3 | - CoI: 5<br>- Transactional distance: 0 |

## Round 3

<u>Revised Process:</u>

In this round, two additional checkpoints (#18th and #25th) were added between the 12th and 33rd papers, to check if richer challenges and strategies could be generated after analysing fewer papers.

To address the issue of inaccurate page numbers, the following revised prompt was given to ChatGPT.

> *(Revised Prompt): Give the actual page number shown in the paper, not the page number of PDF files.*

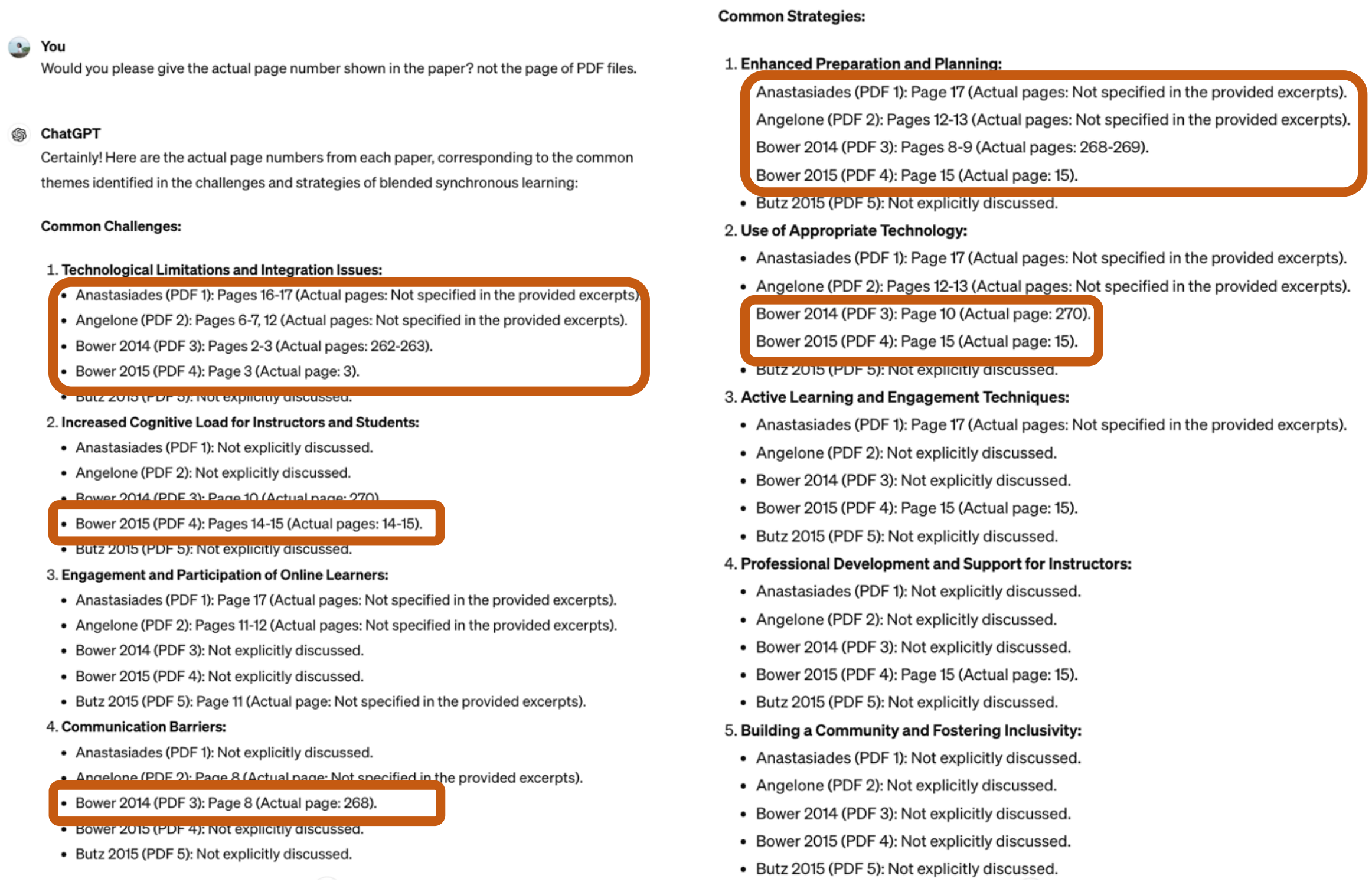

Figure 10. Prompts were given to let GPT show actual page number of a journal

We continued to instruct ChatGPT to analyse common challenges and strategies based on the framework (Learner-learner, Learner-content, Learner-instructor interaction. Moreover, we also asked GPT to provide examples for illustrating the results of each theme. For instance, the new prompt was as follows:

> *(New Prompt): For the challenges: Technological Limitations and Interference, can you give me 1-2 examples from the 33 papers you read before?.*

<u>Results and Analysis:</u>

As shown in Figure 11, more common challenges and strategies were generated after analysing the 18th and 25th papers.

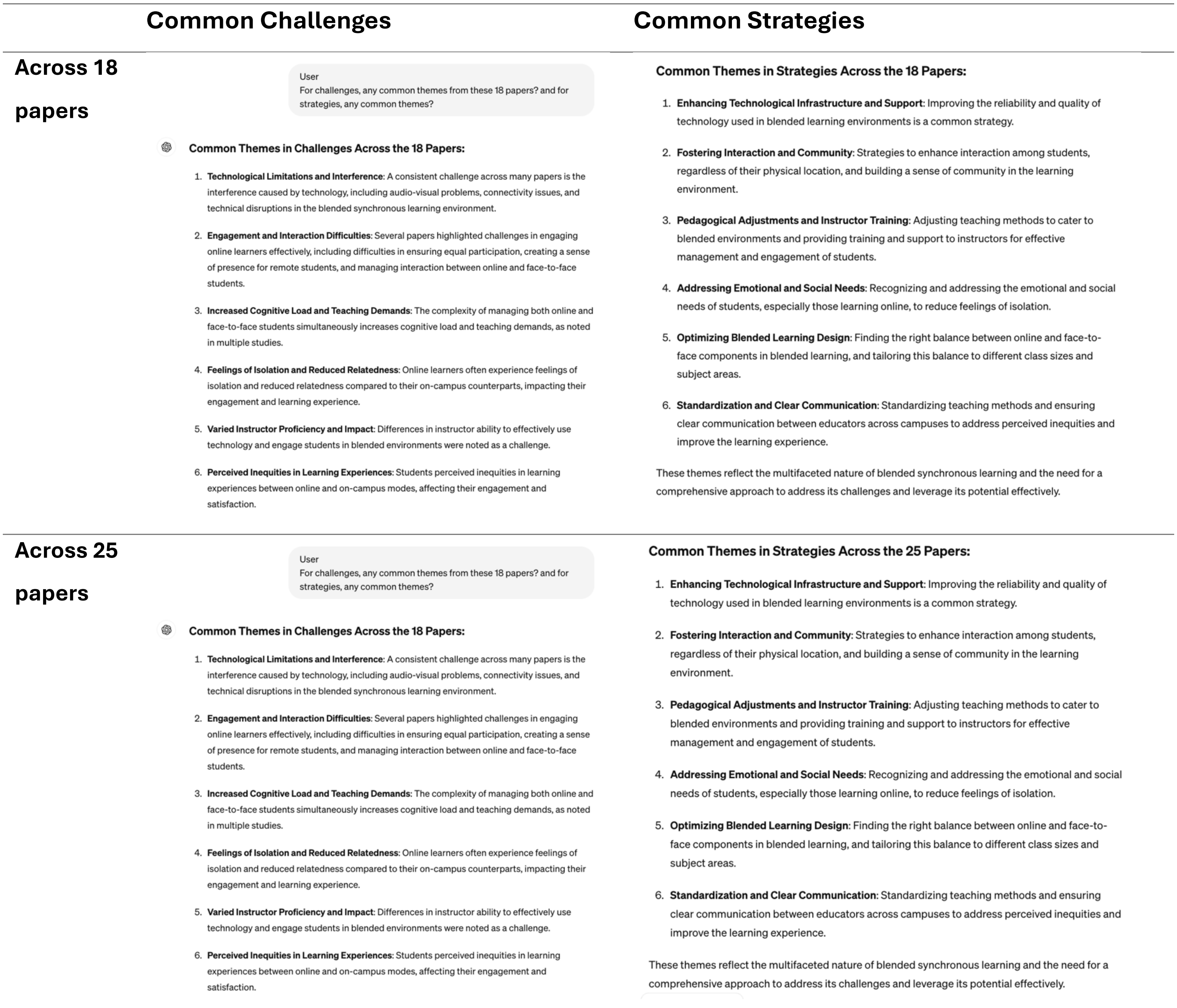

| | Common Challenges | Common Strategies |
|---|---|---|
| Across 18 papers | User<br>For challenges, any common themes from these 18 papers? and for strategies, any common themes?<br><br>**Common Themes in Challenges Across the 18 Papers:**<br>1. **Technological Limitations and Interference**: A consistent challenge across many papers is the interference caused by technology, including audio-visual problems, connectivity issues, and technical disruptions in the blended synchronous learning environment.<br>2. **Engagement and Interaction Difficulties**: Several papers highlighted challenges in engaging online learners effectively, including difficulties in ensuring equal participation, creating a sense of presence for remote students, and managing interaction between online and face-to-face students.<br>3. **Increased Cognitive Load and Teaching Demands**: The complexity of managing both online and face-to-face students simultaneously increases cognitive load and teaching demands, as noted in multiple studies.<br>4. **Feelings of Isolation and Reduced Relatedness**: Online learners often experience feelings of isolation and reduced relatedness compared to their on-campus counterparts, impacting their engagement and learning experience.<br>5. **Varied Instructor Proficiency and Impact**: Differences in instructor ability to effectively use technology and engage students in blended environments were noted as a challenge.<br>6. **Perceived Inequities in Learning Experiences**: Students perceived inequities in learning experiences between online and on-campus modes, affecting their engagement and satisfaction. | **Common Themes in Strategies Across the 18 Papers:**<br>1. **Enhancing Technological Infrastructure and Support**: Improving the reliability and quality of technology used in blended learning environments is a common strategy.<br>2. **Fostering Interaction and Community**: Strategies to enhance interaction among students, regardless of their physical location, and building a sense of community in the learning environment.<br>3. **Pedagogical Adjustments and Instructor Training**: Adjusting teaching methods to cater to blended environments and providing training and support to instructors for effective management and engagement of students.<br>4. **Addressing Emotional and Social Needs**: Recognizing and addressing the emotional and social needs of students, especially those learning online, to reduce feelings of isolation.<br>5. **Optimizing Blended Learning Design**: Finding the right balance between online and face-to-face components in blended learning, and tailoring this balance to different class sizes and subject areas.<br>6. **Standardization and Clear Communication**: Standardizing teaching methods and ensuring clear communication between educators across campuses to address perceived inequities and improve the learning experience.<br><br>These themes reflect the multifaceted nature of blended synchronous learning and the need for a comprehensive approach to address its challenges and leverage its potential effectively. |
| Across 25 papers | User<br>For challenges, any common themes from these 18 papers? and for strategies, any common themes?<br><br>**Common Themes in Challenges Across the 18 Papers:**<br>1. **Technological Limitations and Interference**: A consistent challenge across many papers is the interference caused by technology, including audio-visual problems, connectivity issues, and technical disruptions in the blended synchronous learning environment.<br>2. **Engagement and Interaction Difficulties**: Several papers highlighted challenges in engaging online learners effectively, including difficulties in ensuring equal participation, creating a sense of presence for remote students, and managing interaction between online and face-to-face students.<br>3. **Increased Cognitive Load and Teaching Demands**: The complexity of managing both online and face-to-face students simultaneously increases cognitive load and teaching demands, as noted in multiple studies.<br>4. **Feelings of Isolation and Reduced Relatedness**: Online learners often experience feelings of isolation and reduced relatedness compared to their on-campus counterparts, impacting their engagement and learning experience.<br>5. **Varied Instructor Proficiency and Impact**: Differences in instructor ability to effectively use technology and engage students in blended environments were noted as a challenge.<br>6. **Perceived Inequities in Learning Experiences**: Students perceived inequities in learning experiences between online and on-campus modes, affecting their engagement and satisfaction. | **Common Themes in Strategies Across the 25 Papers:**<br>1. **Enhancing Technological Infrastructure and Support**: Improving the reliability and quality of technology used in blended learning environments is a common strategy.<br>2. **Fostering Interaction and Community**: Strategies to enhance interaction among students, regardless of their physical location, and building a sense of community in the learning environment.<br>3. **Pedagogical Adjustments and Instructor Training**: Adjusting teaching methods to cater to blended environments and providing training and support to instructors for effective management and engagement of students.<br>4. **Addressing Emotional and Social Needs**: Recognizing and addressing the emotional and social needs of students, especially those learning online, to reduce feelings of isolation.<br>5. **Optimizing Blended Learning Design**: Finding the right balance between online and face-to-face components in blended learning, and tailoring this balance to different class sizes and subject areas.<br>6. **Standardization and Clear Communication**: Standardizing teaching methods and ensuring clear communication between educators across campuses to address perceived inequities and improve the learning experience.<br><br>These themes reflect the multifaceted nature of blended synchronous learning and the need for a comprehensive approach to address its challenges and leverage its potential effectively. |

Figure 11. ChatGPT's results across 18 papers and 25 papers

In addition, the final review results after analysing 33 papers, as shown in Figure 12, were highly consistent with that of the OLR.

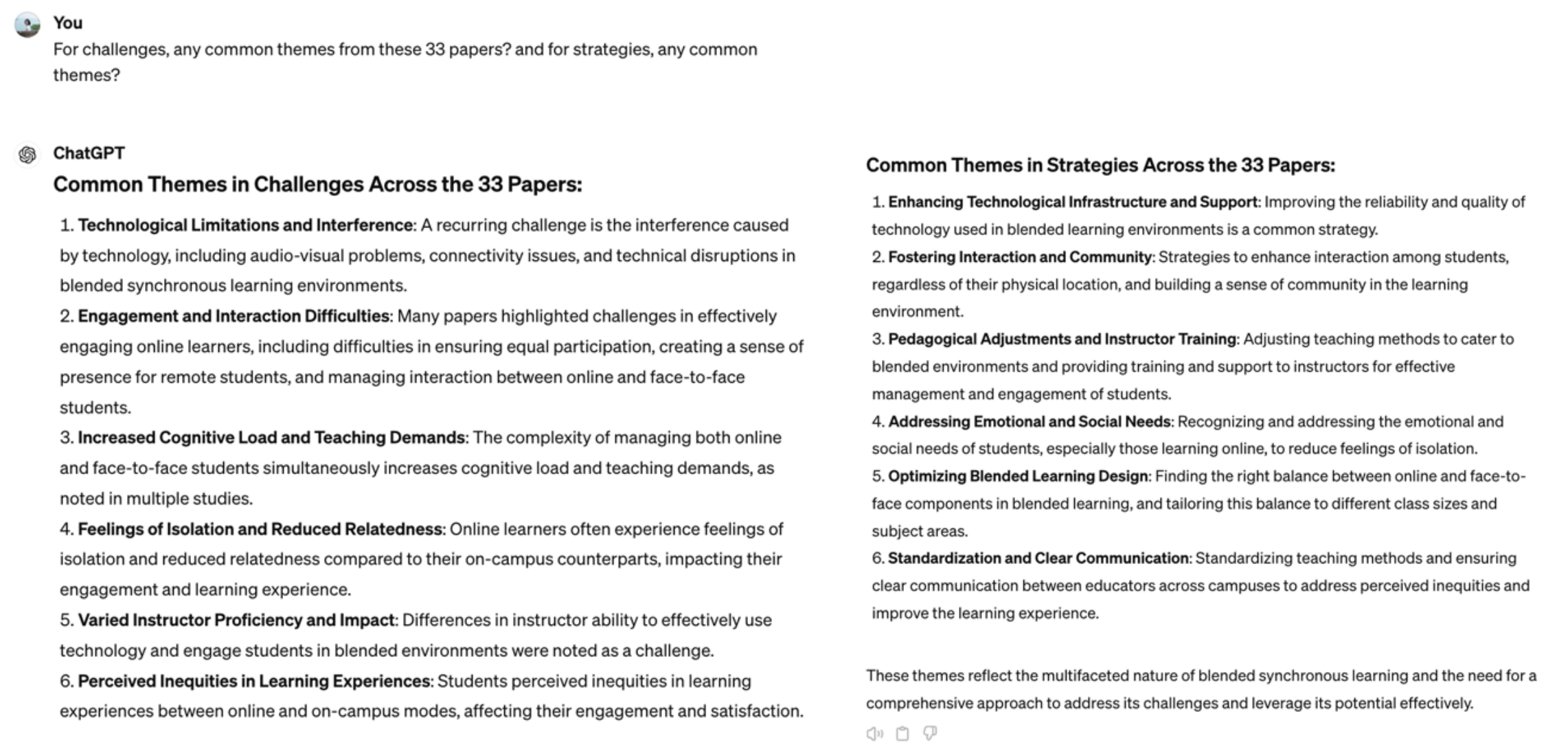
You
For challenges, any common themes from these 33 papers? and for strategies, any common themes?

ChatGPT
**Common Themes in Challenges Across the 33 Papers:**

1. **Technological Limitations and Interference**: A recurring challenge is the interference caused by technology, including audio-visual problems, connectivity issues, and technical disruptions in blended synchronous learning environments.
2. **Engagement and Interaction Difficulties**: Many papers highlighted challenges in effectively engaging online learners, including difficulties in ensuring equal participation, creating a sense of presence for remote students, and managing interaction between online and face-to-face students.
3. **Increased Cognitive Load and Teaching Demands**: The complexity of managing both online and face-to-face students simultaneously increases cognitive load and teaching demands, as noted in multiple studies.
4. **Feelings of Isolation and Reduced Relatedness**: Online learners often experience feelings of isolation and reduced relatedness compared to their on-campus counterparts, impacting their engagement and learning experience.
5. **Varied Instructor Proficiency and Impact**: Differences in instructor ability to effectively use technology and engage students in blended environments were noted as a challenge.
6. **Perceived Inequities in Learning Experiences**: Students perceived inequities in learning experiences between online and on-campus modes, affecting their engagement and satisfaction.

**Common Themes in Strategies Across the 33 Papers:**

1. **Enhancing Technological Infrastructure and Support**: Improving the reliability and quality of technology used in blended learning environments is a common strategy.
2. **Fostering Interaction and Community**: Strategies to enhance interaction among students, regardless of their physical location, and building a sense of community in the learning environment.
3. **Pedagogical Adjustments and Instructor Training**: Adjusting teaching methods to cater to blended environments and providing training and support to instructors for effective management and engagement of students.
4. **Addressing Emotional and Social Needs**: Recognizing and addressing the emotional and social needs of students, especially those learning online, to reduce feelings of isolation.
5. **Optimizing Blended Learning Design**: Finding the right balance between online and face-to-face components in blended learning, and tailoring this balance to different class sizes and subject areas.
6. **Standardization and Clear Communication**: Standardizing teaching methods and ensuring clear communication between educators across campuses to address perceived inequities and improve the learning experience.

These themes reflect the multifaceted nature of blended synchronous learning and the need for a comprehensive approach to address its challenges and leverage its potential effectively.

Figure 12: Common themes of 33 papers

Moreover, ChatGPT provided concrete examples to illustrate the summarized themes as shown in Figure 12.

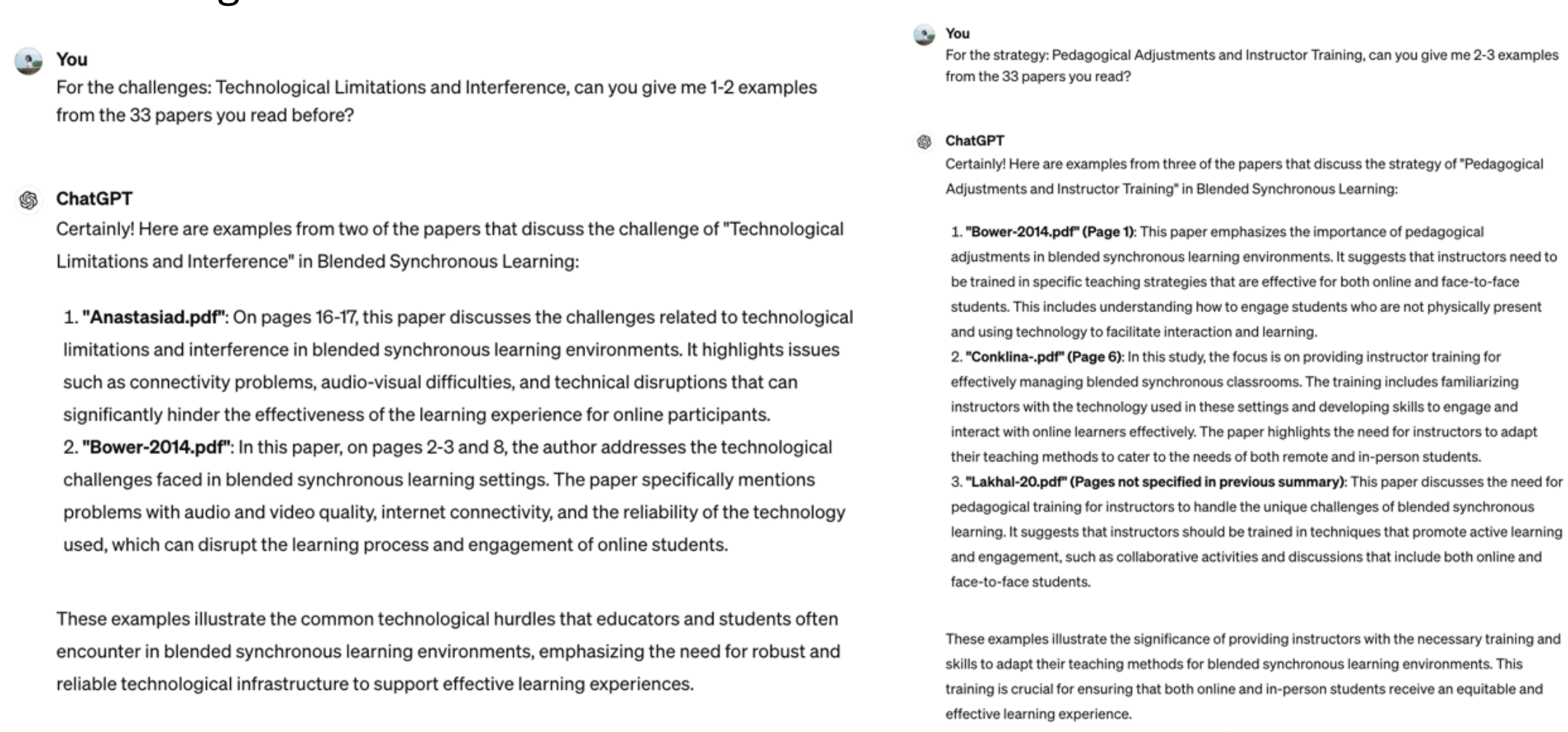
You
For the challenges: Technological Limitations and Interference, can you give me 1-2 examples from the 33 papers you read before?

ChatGPT
Certainly! Here are examples from two of the papers that discuss the challenge of "Technological Limitations and Interference" in Blended Synchronous Learning:

1. **"Anastasiad.pdf"**: On pages 16-17, this paper discusses the challenges related to technological limitations and interference in blended synchronous learning environments. It highlights issues such as connectivity problems, audio-visual difficulties, and technical disruptions that can significantly hinder the effectiveness of the learning experience for online participants.
2. **"Bower-2014.pdf"**: In this paper, on pages 2-3 and 8, the author addresses the technological challenges faced in blended synchronous learning settings. The paper specifically mentions problems with audio and video quality, internet connectivity, and the reliability of the technology used, which can disrupt the learning process and engagement of online students.

These examples illustrate the common technological hurdles that educators and students often encounter in blended synchronous learning environments, emphasizing the need for robust and reliable technological infrastructure to support effective learning experiences.

You
For the strategy: Pedagogical Adjustments and Instructor Training, can you give me 2-3 examples from the 33 papers you read?

ChatGPT
Certainly! Here are examples from three of the papers that discuss the strategy of "Pedagogical Adjustments and Instructor Training" in Blended Synchronous Learning:

1. **"Bower-2014.pdf" (Page 1)**: This paper emphasizes the importance of pedagogical adjustments in blended synchronous learning environments. It suggests that instructors need to be trained in specific teaching strategies that are effective for both online and face-to-face students. This includes understanding how to engage students who are not physically present and using technology to facilitate interaction and learning.
2. **"Conklina-.pdf" (Page 6)**: In this study, the focus is on providing instructor training for effectively managing blended synchronous classrooms. The training includes familiarizing instructors with the technology used in these settings and developing skills to engage and interact with online learners effectively. The paper highlights the need for instructors to adapt their teaching methods to cater to the needs of both remote and in-person students.
3. **"Lakhal-20.pdf" (Pages not specified in previous summary)**: This paper discusses the need for pedagogical training for instructors to handle the unique challenges of blended synchronous learning. It suggests that instructors should be trained in techniques that promote active learning and engagement, such as collaborative activities and discussions that include both online and face-to-face students.

These examples illustrate the significance of providing instructors with the necessary training and skills to adapt their teaching methods for blended synchronous learning environments. This training is crucial for ensuring that both online and in-person students receive an equitable and effective learning experience.

Figure 13: Prompts were given to let GPT give examples of each theme

But the paper number was still inaccurate. For instance, “Conklina.pdf” (Page 6) is Page 22 in the journal.

## Discussion

The main purpose of this exploratory study was to explore how to use ChatGPT to generate reliable systematic literature review results. In this section, key findings from the study are discussed and design principles are summarised.

This study confirms that ChatGPT can be a helpful tool for doing the SLR. It could rapidly analyse papers, extract relevant information, and summarise main findings into common themes. It could trace back to the content mentioned in each paper very quickly. It could provide direct quotes or examples. From this perspective, ChatGPT can function as a researcher assistant to help conduct a tentative literature review on a research topic.

However, we also identified certain limitations when ChatGPT was used to do the systematic literature review. For instance, the reviewed papers were screened by the researchers.  Presently ChatGPT has no access to academic databases, and cannot screen out papers that meet the criteria (Dwivedi et al., 2023). It still requires human researchers to identify suitable papers for analysis.  Furthermore, ChatGPT cannot accurately identify exact page numbers from a PDF document. The page numbers ChatGPT identified were based on a single PDF document starting from Page 1, rather than the actual page of a journal paper. Researchers need to check the actual page number from the journal papers in person as references.

**Guiding Principles**

Based on the format suggested by van den Akker (1999), the guiding principles summarized from this study are presented as follows:

*If you want to conduct a rigorous and effective systematic literature review using GPT, you are best advised to:*

- screen the reviewed papers first, for instance, using the Preferred Reporting Items for Systematic Reviews and Meta-Analyses (PRISMA) protocol (Moher et al., 2009).

- provide highly detailed and precise instructions, specifying exactly what to do in each step of the process.

- define research questions and specify a theoretical or analytical framework for ChatGPT to complete a targeted and accurate SLR.

- probide a page number range or specific portion/section of the paper due to its limitation in specifically identifying whether related information are from the "Literature" or from the "Findings".

- check page numbers in person.

- Document the research process to avoid loss of records, as well as to record the prompts.

- let ChatGPT read papers repeatedly, and triangulate with human beings. In this way, ChatGPT can become research assistants to enhance the efficiency and accuracy.

## Conclusion

This study provides an empirical example of using ChatGPT-4.0 to do a systematic literature review. This paper describes how to increase the reliability of the review result after three rounds of analysis and improvement. The result shows that ChatGPT can conduct the SLR. However, it has limitations. In general, ChatGPT can serve as a Research Assistant(s) in helping with reading, analyzing and summarizing existing studies and the generated results can be used as a reference to allow researchers to have a quick understanding of the research in a certain area. The design principles summarized from this study can guide researchers in generating reliable review results.

## Reference

Arif, T. B., Munaf, U., & Ul-Haque, I. (2023). The future of medical education and research: Is ChatGPT a blessing or blight in disguise? *Medical Education Online, 28*(1), 2181052. https://doi.org/10.1080/10872981.2023.2181052

Carrera-Rivera, A., Ochoa, W., Larrinaga, F., & Lasa, G. (2022). How-to conduct a systematic literature review: A quick guide for computer science research. *MethodsX*, 9, 101895–101895. https://doi.org/10.1016/j.mex.2022.101895

Choi, E., Lee, J. J., Ho, M.-H., Kwok, J., & Lok, K. (2023). Chatting or cheating? The impacts of ChatGPT and other artificial intelligence language models on nurse education. *Nurse Education Today, 125*. https://doi.org/10.1016/j.nedt.2023.105796

Cohen, L., Manion, L., & Morrison, K. (2018). *Research methods in education: Eighth edition*. Routledge.

Cotton, D. R. E., Cotton, P. A., & Shipway, J. R. (2023). Chatting and cheating: Ensuring academic integrity in the era of ChatGPT. Innovations in Education and Teaching International, ahead-of- print(ahead-of-print), 1–12. https://doi.org/10.1080/14703297.2023.2190148

da Silva, J. A. T. (2023). Is ChatGPT a valid author? *Nurse Education in Practice, 68*, 103600. https://doi.org/10.1016/j.nepr.2023.103600

Dwivedi Y.K. et al. (2023). Opinion Paper: “So what if ChatGPT wrote it?” Multidisciplinary perspectives on opportunities, challenges and implications of generative conversational AI for research, practice and policy*, International Journal of Information Management*, Volume 71, 2023, 102642, https://doi.org/10.1016/j.ijinfomgt.2023.102642

Eysenbach, G. (2023). The role of ChatGPT, generative language models, and artificial intelligence in medical education: A conversation with ChatGPT and a call for papers. *JMIR Medical Education, 9*. https://doi.org/10.2196/46885

Gough, D., Oliver, S., & Thomas, J. (2017). An Introduction to Systematic Reviews. An accessible guide to the principles and practices of systematic reviewing

Hwang, G.J. & Chen, N.S. (2023). Exploring the Potential of Generative Artificial Intelligence in Education: Applications, Challenges, and Future Research Directions. *Educational Technology & Society*, 26(2). https://doi.org/10.30191/ETS.202304_26(2).0014

Haman, M., & Školník, M. (2023). Using ChatGPT to conduct a literature review. Accountability in Research, ahead-of-print(ahead-of-print), 1–3. https://doi.org/10.1080/08989621.2023.2185514

Jahic, I., Ebner, M., & Schön, S. (2023). Harnessing the power of artificial intelligence and ChatGPT in education – a first rapid literature review. In T. Bastiaens (Ed.), *Proceedings of EdMedia + Innovate Learning 2023* (pp. 1462-1470). Association for the Advancement of Computing in Education. https://www.learntechlib.org/p/222670

Kasneci, E., Sessler, K., Küchemann, S., Bannert, M., Dementieva, D., Fischer, F., Gasser, U., Groh, G., Günnemann, S., Hüllermeier, E., Krusche, S. Kutyniok, G., Michaeli, T., Nerdel, C., Pfeffer, J., Poquet, O., Sailer, M., Schmidt, A., Seidel, T., … _Kasneci, G. (2023). ChatGPT for good? On opportunities and challenges of large language models for education. *Learning and Individual Differences, 103*, 102274. https://doi.org/10.1016/j.lindif.2023.102274

Kitchenham, B., Pearl Brereton, O., Budgen, D., Turner, M., Bailey, J., & Linkman, S. (2009). Systematic literature reviews in software engineering - A systematic literature review. *Information and Software Technology*, 51(1), 7–15. https://doi.org/10.1016/j.infsof.2008.09.009

Kooli, C. (2023). Chatbots in Education and Research: A Critical Examination of Ethical Implications and Solutions. *Sustainability (Switzerland)*, *15*(7). https://doi.org/10.3390/su15075614

Kuhail, Mohammad Amin & Alturki, Nazik & Alramlawi, Salwa & Alhejori, Kholood. (2022). Interacting with educational chatbots: A systematic review. Education and Information Technologies. 28. 1-46. 10.1007/s10639-022-11177-3.

Moher, D., Liberati, A., Tetzlaff, J., Altman, D.G., & The PRISMA Group. (2009). Preferred Reporting Items for Systematic Reviews and Meta-Analyses: The PRISMA Statement. BMJ, 339. A critical resource for improving the reporting of systematic reviews and meta-analysis

Qi, X., Zhu, Z., & Wu, B. (2023). The promise and peril of ChatGPT in geriatric nursing education: What we know and do not know. *Aging and Health Research*, 3(2), 100136. https://doi.org/10.1016/j.ahr.2023.100136

Shoufan, A. (2023). Exploring students' perceptions of ChatGPT: Thematic analysis and follow-up survey. IEEE Access, 11, 38805–38818. https://doi.org/10.1109/ACCESS.2023.3268224

Silverman, D. (2021). *Qualitative research*, London: SAGE

Thakur, A., Devanshi, P., & Thakur, A. (2023). ChatGPT in nursing education: Is there a role for curriculum development? *Teaching and Learning in Nursing*, 18 (3), 450-451, https://doi.org/10.1016/j.teln.2023.03.011

van den Akker, J. (1999). Principles and methods of development research. In J. van den Akker, R. M. Branch, K. Gustafson, N. Nieveen, & T. Plomp (Eds.), *Design approaches and tools in education and training (pp. 1–14)*. Dordrecht, Netherlands: Kluwer Academic. doi:10.1007/ 978-94-011-4255-7

Wang, Q., & Huang, Q. (2023). Engaging Online Learners in Blended Synchronous Learning: A Systematic Literature Review. *IEEE Transactions on Learning Technologies*, 17, 1–15. https://doi.org/10.1109/TLT.2023.3282278

Waseem, M., Ahmad, A., Liang, P. & Fahmideh, M. (2023), *Conducting systematic literature reviews with ChatGPT*, retrieved from: https://www.researchgate.net/publication/369062219_Conducting_Systematic_Literature_Reviews_with_ChatGPT_ChatGPT_for_SLRs_A_Proposal

Wiki, ChatGPT, retrieved from https://en.wikipedia.org/wiki/ChatGPT

Wu, T., He, S.Z., Liu, J.P., Sun, S.Q., Liu, K., Han, Q.L., & Tang, Y. (2023)., A Brief Overview of ChatGPT: The History, Status Quo and Potential Future Development, *IEEE/CAA Journal of Automatica Sinica*, 10(5), 1122-1136, May 2023, doi: 10.1109/JAS.2023.